\pdfoutput=1
\documentclass[11pt]{article}

\usepackage[final]{acl}

\usepackage{times}
\usepackage{latexsym}

\usepackage[T1]{fontenc}
\usepackage[utf8]{inputenc}

\usepackage{microtype}

\usepackage{inconsolata}

\usepackage{graphicx}

\usepackage{booktabs}
\usepackage{color}
\usepackage{colortbl}
\usepackage{multirow}
\usepackage{stfloats}
\usepackage[dvipsnames]{xcolor}
   
\usepackage{amssymb}
\usepackage{soul}
\newcommand{\ctext}[3][RGB]{%
  \begingroup
  \definecolor{hlcolor}{#1}{#2}\sethlcolor{hlcolor}%
  \hl{#3}%
  \endgroup
}

\title{Does Biomedical Training Lead to Better Medical Performance?}

\author{ Amin Dada$^{1}$ \hspace{0.05em} Osman Alperen Koraş
$^{1}$  \hspace{0.05em} Marie Bauer$^{1}$  Jean-Philippe Corbeil$^{2}$ \\   \textbf{Amanda Butler Contreras$^3$} \hspace{0.05em} \textbf{Constantin Marc Seibold$^{1}$} \hspace{0.05em} \textbf{Kaleb E Smith$^{3}$}   \\ \textbf{Julian Friedrich$^{1}$} \hspace{0.05em} \textbf{Jens Kleesiek}$^{1}$\thanks{Other affiliations:
 Cancer Research Center Cologne Essen (CCCE), German Cancer Consortium (DKTK, Partner site Essen) and Department of Physics of TU Dortmund (Dortmund, Germany).}\\
        $^1$Institute for AI in Medicine, University Hospital Essen, Germany \\
        $^2$ Microsoft Healthcare \& Life Sciences \quad
        $^3$ NVIDIA \\
        }
\begin{document}
\maketitle
\begin{abstract}
Large Language Models (LLMs) hold significant potential for improving healthcare applications, with biomedically adapted models promising enhanced performance on medical tasks. However, the effectiveness of biomedical domain adaptation for clinical tasks remains uncertain. In this study, we conduct a direct comparison of 12 biomedically adapted models and their general-domain base counterparts across six clinical tasks. Our results reveal that 11 out of 12 biomedical models exhibit performance declines, challenging prior findings that reported positive effects of biomedical adaptation. Notably, previous positive results primarily relied on multiple-choice evaluations, which may not reflect performance in real-world clinical applications. To promote reproducibility and further research, we open-source our evaluation pipeline, providing a resource for the development of models with practical benefits in healthcare settings.

\end{abstract}

\section{Introduction}

Large Language Models (LLMs) have the potential to transform healthcare by enhancing patient care quality and efficiency \cite{moor2023foundation}. Open-source biomedical LLMs, designed for medical applications, promise improved performance with fewer parameters than general models \citep{luoBioMedGPTOpenMultimodal2023, chenMEDITRON70BScalingMedical2023, labrakBioMistralCollectionOpenSource2024a}. However, recent research questions the effectiveness of biomedical domain adaptation \citep{jeong-etal-2024-medical, ceballos-arroyo-etal-2024-open, dada2025medisumqa}. 
\definecolor{lightGray}{rgb}{0.9,0.9,0.9}
\definecolor{lightGreen}{rgb}{0.725, 0.969, 0.647}
\definecolor{lightRed}{rgb}{0.988, 0.694, 0.596}
\begin{table*}[ht]
\setlength{\tabcolsep}{2pt} % General space between cols (6pt standard)
    \renewcommand{\arraystretch}{1} 
\centering
\small
\begin{tabular}{lcclll}
\toprule
\textbf{Dataset} & \textbf{Samples} & \textbf{Words} & \textbf{Documents} & \textbf{Focus}\\
\specialrule{0.8pt}{0pt}{0pt} 
\rowcolor{lightGray}
\multicolumn{5}{c}{\scriptsize \textbf{Level 1}} \\
MedNLI & 1425 & 21 & \small Clinical Notes  & \small Clinical reasoning  \\
MeQSum & 1000 & 61 & \small Consumer Health Questions  & \small Summarization  \\
Problem Summary & 237 & 124 & \small Clinical Notes  & \small Information extraction   \\
\rowcolor{lightGray}
\multicolumn{5}{c}{\scriptsize \textbf{Level 2}} \\
LongHealth &  400 &  5537 & \small EHR   & \small Information extraction  \\
MeDiSumQA & 453 & 1452 & \small Discharge Summary  & \small Simplification/Clinical reasoning    \\
MeDiSumCode & 500 &  1515 & \small Discharge Summary & \small Information extraction / Coding   \\
\bottomrule
\end{tabular}
\caption{An overview of the characteristics of the tasks. We split the tasks into the difficulties level 1 and level 2.}
\label{tab:task_overview}
\end{table*}

In this study we perform a direct comparison of $12$ biomedically adapted models with their general-domain base models on six clinical tasks. Our results reveal performance declines in 11 of 12 biomedical models. This is in contrast to previous findings that reported positive effects of biomedical training \citep{chenMEDITRON70BScalingMedical2023,  gururajan2024aloe, christophe2024med42}. However, these studies primarily relied on multiple-choice evaluations that did not incorporate real-world clinical documents. This suggests that the observed benefits of biomedical adaptation may not translate effectively to practical healthcare settings.

To facilitate reproducibility and enable future development of models with practical benefits in healthcare settings, we open-source our evaluation pipeline. By providing a standardized framework for assessing biomedical LLMs on real-world clinical tasks, we aim to bridge the gap between benchmark performance and real-world applicability.

\section{Related Work}

The need for specialized healthcare tools has recently accelerated biomedical LLM development, yielding commercial models like Med-PaLM \citep{singhal2023large} and MedGemini \citep{saab2024capabilities}, and open-source alternatives such as Meditron \citep{chenMEDITRON70BScalingMedical2023}, Biomistral \citep{labrakBioMistralCollectionOpenSource2024a}, Internist.ai \citep{internist}, and Med42 \citep{christophe2024med42}.

Although biomedical LLMs initially outperformed general-domain models on tasks like multiple-choice question-answering (MCQA) exams, recent studies \citep{jeong-etal-2024-medical, ceballos-arroyo-etal-2024-open, dada2025medisumqa} challenge this view. \citet{jeong-etal-2024-medical} found no clear advantage for biomedical LLMs with model-specific prompt tuning, and \citet{ceballos-arroyo-etal-2024-open} suggest domain adaptation might impair instruction-following.

\section{Evaluation Tasks}

We introduce the clinical language understanding evaluation (CLUE) consisting of six tasks on clinical notes, consumer health questions, electronic health records (EHR) and discharge summaries, encompassing information extraction, summarization, clinical reasoning, simplification, and coding. Table \ref{tab:task_overview} summarizes the characteristics of these tasks. We divide the tasks into two levels. Level 1 includes simpler tasks with short inputs, while Level 2 has complex tasks with long inputs. We provide prompt examples for each task in Figures \ref{meqsum_example_prompts}, \ref{probsum_example_prompts}, \ref{mednli_example_prompts}, \ref{longhealth_example_prompts}, \ref{medisumqa_example_prompts}, and
\ref{medisumcode_example_prompts} in Appendix \ref{appendix:prompting}.

\textbf{MedNLI} \citep{DBLP:journals/corr/abs-1808-06752} is based on clinical notes from MIMIC-III \citep{johnson2016mimic}. It evaluates models on predicting the logical relationship—contradiction, neutrality, or entailment—between a premise and hypotheses, testing clinical reasoning with short input lengths.

\textbf{MeQSum} \citep{MeQSum} contains 1,000 consumer health inquiries summarized by medical experts. This task evaluates whether models can understand lay language, extract key information, and reformulate patient queries into concise, medically sound questions.

\textbf{Problem Summary} Derived from SOAP-structured clinical notes, this task was first described by \citet{gaoSummarizingPatientsProblems2022} and utilizes the Subjective and Assessment sections for predicting a patient’s health problems \citep{Weed1964}. Like MedNLI, its short input length tests basic information extraction abilities.

\textbf{LongHealth} \citep{adamsLongHealthQuestionAnswering2024} consists of 20 fictional patient records designed to challenge LLMs on long input comprehension. Evaluation involves answering questions on multiple long documents, handling added irrelevant information, and recognizing when data is unavailable. This task assesses comprehension, long-input retention, and hallucination tendencies.

\textbf{MeDiSumQA} \citep{dada2025medisumqa} requires models to comprehend MIMIC-IV \citep{mimiciv_v1} discharge summaries, extract key information, answer patient-related queries, and simplify medical information. Additionally, models must apply medical knowledge to provide appropriate follow-up advice.

Using MIMIC-IV, we create \textbf{MeDiSumCode}, an ICD-10 prediction dataset by linking discharge summaries with annotated ICD-10 codes via hospital admission IDs. This dataset provides discharge summaries as inputs and ICD-10 codes as labels for model evaluation.

\textbf{MeDiSumCode} involves assigning ICD-10 codes to diagnoses and procedures in discharge summaries, a critical task for patient records, billing, and healthcare analysis \citep{ICD}. This challenge requires models to extract diagnoses from complex clinical text, comprehend over 70,000 ICD-10 codes, and accurately match diagnoses to the correct codes.

\section{Experimental setup}

We evaluated $24$ language models, including biomedically trained models, their base models, and additional general-domain models as reference. Our evaluation aims to (1) measure the effects of continuous biomedical training, (2) assess whether biomedical models or general-domain models are more suitable for specific medical scenarios, and (3) rank current openly available models. Appendix \ref{appendix:metrics} describes the metrics we applied to each task. For each task, we report the average over all metrics.

\subsection{Models}
We evaluate the following biomedical LLMs:  Meditron-7B and 70B \citep{chenMEDITRON70BScalingMedical2023}, Internist.ai \citep{internist}, BioMistral \citep{labrakBioMistralCollectionOpenSource2024a}, Llama3-Aloe-8B-Alpha \citep{gururajan2024aloe}, Llama3-OpenBioLLM-8B and 70B \citep{OpenBioLLMs}, Med42-Llama3-8B and 70B \citep{christophe2024med42}, and Meditron3-8B and 70B \citep{meditron3_modelcard}. More details are in Table \ref{biomed_models} in Appendix \ref{appendix:models}. We did not evaluate Llama2-based models on Level 2 tasks due to their limited context size of 4k tokens.

We also evaluate the base models of the biomedical LLMs and the following additional models: Zephyr-7B-Beta \cite{tunstall2023zephyr}, Mistral-7B-Instruct-v0.2 \cite{jiang2023mistral}, Phi-3-Mini-128k-Instruct \cite{abdin2024phi}, Mixtral-8x7B, and Mixtral-8x22B \cite{jiang2024mixtral}.

% \begin{figure*}[t]
%   \centering
%   \includegraphics[width=\textwidth]{images/results_bar.pdf}
%   \caption{The relative difference between biomedical models and their base model.}
%   \label{results_bar}
% \end{figure*}

\section{Results}

\begin{table*}[!htp]\centering
\small
\resizebox{1 \textwidth}{!}{
\begin{tabular}
{ll@{\hspace{0.2em}}ll@{\hspace{0.2em}}ll@{\hspace{0.2em}}ll@{\hspace{0.2em}}ll@{\hspace{0.2em}}ll@{\hspace{0.2em}}l}\toprule
& \multicolumn{6}{c}{Level 1} & \multicolumn{6}{c}{Level 2} \\
\cmidrule(lr){2-7} \cmidrule(lr){8-13}  Model &\multicolumn{2}{l}{MedNLI} &\multicolumn{2}{l}{Prob. Sum.} &\multicolumn{2}{l}{MeQSum} &\multicolumn{2}{l}{LongHealth} &\multicolumn{2}{l}{MeDiSumQA} &\multicolumn{2}{l}{MeDiSumCode} \\\midrule
Llama-2-7B &29.5 & &16.8 & &14.0 & & - & & - &  & - & \\
\quad - Meditron-7B &2.4 &{\color[HTML]{C20000} (-27.1)} &21.6 &{\color[HTML]{02b802} (+4.8)} &15.1 &{\color[HTML]{02b802} (+1.1)} & - & & - &  & - & \\

Llama-2-70B &76.3 & &18.6 & &10.6 & & - & & - &  & - & \\
\quad - Meditron-70B &63.5 &{\color[HTML]{C20000} (-12.7)} &18.7 &{\color[HTML]{02b802} (+0.1)} &9.6 &{\color[HTML]{C20000} (-1.1)} & - & & - &  & - & \\
Mistral-7B-Instruct-v0.1 &64.8 & &25.0 & &31.1 & &30.0 & &25.5 & &13.9 & \\
\quad - BioMistral-7B &62.8 &{\color[HTML]{C20000} (-2.0)} &25.1 &{\color[HTML]{02b802} (+0.1)} &33.9 &{\color[HTML]{02b802} (+2.8)} &26.7 &{\color[HTML]{C20000} (-3.3)} &22.8 &{\color[HTML]{C20000} (-2.7)} &22.0 &{\color[HTML]{02b802} (+8.2)} \\
\quad - BioMistral-7B-DARE &66.8 &{\color[HTML]{02b802} (+2.0)} &28.4 &{\color[HTML]{02b802} (+3.4)} &34.5 &{\color[HTML]{02b802} (+3.4)} &30.5 &{\color[HTML]{02b802} (+0.5)} &25.7 &{\color[HTML]{02b802} (+0.2)} &21.3 &{\color[HTML]{02b802} (+7.4)} \\
\quad - Internist.ai 7b &76.3 &{\color[HTML]{02b802} (+11.5)} &23.1 &{\color[HTML]{C20000} (-1.9)} &15.2 &{\color[HTML]{C20000} (-15.9)} &44.2 &{\color[HTML]{02b802} (+14.2)} &19.8 &{\color[HTML]{C20000} (-5.6)} &21.9 &{\color[HTML]{02b802} (+8.0)} \\
Zephyr 7B &68.5 & &25.5 & &34.2 & &33.3 & &22.7 & &28.5 & \\
Meta-Llama-3-8B-Instruct &74.1 & &31.6 & &39.5 & &58.8 & &30.3 & &27.8 & \\
\quad - OpenBioLLM-8B &44.9 &{\color[HTML]{C20000} (-29.1)} &21.7 &{\color[HTML]{C20000} (-9.9)} &33.0 &{\color[HTML]{C20000} (-6.4)} &26.9 &{\color[HTML]{C20000} (-31.9)} &30.4 &{\color[HTML]{02b802} (+0.1)} &18.9 &{\color[HTML]{C20000} (--8.9)} \\
\quad - Med42-8B &77.5 &{\color[HTML]{02b802} (+3.4)} &32.4 &{\color[HTML]{02b802} (+0.8)} &42.8 &{\color[HTML]{02b802} (+3.3)} &57.8 &{\color[HTML]{C20000} (-1.0)}&29.7 &{\color[HTML]{C20000} (-0.6)}&25.2 &{\color[HTML]{C20000} (-2.6)} \\
\quad - Aloe-8B-Alpha &73.9 &{\color[HTML]{C20000} (-0.1)} &21.3 &{\color[HTML]{C20000} (-10.3)} &32.3 &{\color[HTML]{C20000} (-7.2)} &49.7 &{\color[HTML]{C20000} (-9.1)} &21.4 &{\color[HTML]{C20000} (-8.9)} &19.8 &{\color[HTML]{C20000} (-8.0)} \\
Meta-Llama-3-70B-Instruct &79.4 & &34.7 & &43.0 & &83.8 & &33.3 & &50.9 & \\
\quad - OpenBioLLM-70B &80.8 &{\color[HTML]{02b802} (+1.5)} &23.7 &{\color[HTML]{C20000} (-11.0)} &38.1 &{\color[HTML]{C20000} (-4.8)} &72.9 &{\color[HTML]{C20000} (-10.8)} &30.0 &{\color[HTML]{C20000} (-3.3)} &33.8 &{\color[HTML]{C20000} (-17.2)} \\
\quad - Med42-70B &76.1 &{\color[HTML]{C20000} (-3.2)} &24.3 &{\color[HTML]{C20000} (-10.4)} &33.9 &{\color[HTML]{C20000} (-9.0)} &56.4 &{\color[HTML]{C20000} (-27.4)} &24.2 &{\color[HTML]{C20000} (-9.1)} &42.0 &{\color[HTML]{C20000} (-9.0)} \\
Meta-Llama-3.1-8B-Instruct &79.1 & &29.8 & &42.1 & &70.5 & &32.9 & &33.4 & \\
\quad - Meditron3-8B &74.0 &{\color[HTML]{C20000} (-5.1)} &27.9 &{\color[HTML]{C20000} (-1.9)} &40.8 &{\color[HTML]{C20000} (-1.3)} &50.5 &{\color[HTML]{C20000} (-20.0)} &31.1 &{\color[HTML]{C20000} (-1.8)} &10.1 &{\color[HTML]{C20000} (-23.3)} \\
Meta-Llama-3.1-70B-Instruct &84.9 & &34.5 & &43.7 & &87.7 & &32.6 & &52.8 & \\
\quad - Meditron3-70B &82.6 &{\color[HTML]{C20000} (-2.3)} &31.8 &{\color[HTML]{C20000} (-2.7)} &42.1 &{\color[HTML]{C20000} (-1.6)} &67.7 &{\color[HTML]{C20000} (-20.0)} &32.1 &{\color[HTML]{C20000} (-0.5)} &47.7 &{\color[HTML]{C20000} (-5.0)} \\
Mistral-7B-Instruct-v0.2 &69.9 & &29.2 & &40.3 & &57.4 & &29.4 & &30.0 & \\
Phi-3-mini-instruct &66.6 & &28.4 & &36.7 & &45.9 & &25.8 & &41.1 & \\
Mixtral-8x7B-Instruct-v0.1 &80.1 & &18.4 & &13.8 & &58.1 & &28.8 & &40.8 & \\
Mixtral-8x22B-Instruct-v0.1 &76.5 & &27.3 & &39.6 & &79.7 & &30.0 & &43.9 & \\
\bottomrule
\end{tabular}
}
\caption{\setlength{\fboxsep}{0pt}The aggregated average scores over the individual metrics for each task of our evaluation on CLUE. For biomedical models we include performance {\color[HTML]{02b802} gains} and {\color[HTML]{C20000} losses} compared to their respective base model.}\label{tab:overall_results}
\end{table*}

\begin{table}[ht]
\small
\centering
\begin{tabular}{ll|cc}
\rowcolor[HTML]{FFFFFF} 
\toprule
Model& MCQA & Level 1 & Level 2\\
\toprule
\cellcolor{teal!30}MEDITRON-7B                                 & {\color[HTML]{02b802} +6.07}                             & {\color[HTML]{C20000} -7.08}                     & {\color[HTML]{333333} -}                         \\
\cellcolor{teal!30}MEDITRON-70B                                  & {\color[HTML]{02b802} +3.63}                             & {\color[HTML]{C20000} -4.59}                     & {\color[HTML]{333333} -}                         \\
\cellcolor{orange!35}BioMistral-7B                               & {\color[HTML]{02b802} +4.13}                             & {\color[HTML]{02b802} +0.26}                     & {\color[HTML]{02b802} +0.71}                      \\

\cellcolor{orange!35}BioMistral-7B-DARE                               & {\color[HTML]{02b802} +4.57}                             & {\color[HTML]{02b802} +2.93}                     & {\color[HTML]{02b802} +2.7}                      \\
\cellcolor{orange!35}Internist.ai 7b                                             &  \multicolumn{1}{c|}{-}                                  & {\color[HTML]{C20000} -2.07} & {\color[HTML]{02b802} +5.52} \\
\cellcolor{blue!30}OpenBioLLM-8B                                                 & {\color[HTML]{C20000} -0.63}                             & {\color[HTML]{C20000} -15.17}                    & {\color[HTML]{C20000} -13.54}                    \\
\cellcolor{blue!30}OpenBioLLM-70B & {\color[HTML]{02b802} +1.46}                             & {\color[HTML]{C20000} -4.78}                     & {\color[HTML]{C20000} -10.45}                    \\
\cellcolor{blue!30}Med42-8B                              & {\color[HTML]{02b802} +0.47}                             & {\color[HTML]{02b802} +2.51}                     & {\color[HTML]{C20000} -1.4}                      \\
\cellcolor{blue!30}Med42-70B      & {\color[HTML]{02b802} +2.8}                              & {\color[HTML]{C20000} -7.57}                     & {\color[HTML]{C20000} -15.14}                    \\
\cellcolor{blue!30}Aloe-8B-Alpha                          & {\color[HTML]{02b802} +2.21}                             & {\color[HTML]{C20000} -5.87}                     & {\color[HTML]{C20000} -8.67}                     \\
\cellcolor{olive!30}Meditron3-8B      & \multicolumn{1}{c|}{-} & {\color[HTML]{C20000} -2.76}                     & {\color[HTML]{C20000} --15.04}                    \\
\cellcolor{olive!30}Meditron3-70B      & \multicolumn{1}{c|}{-} & {\color[HTML]{C20000} -2.18}                     & {\color[HTML]{C20000} -8.51}                    \\
\bottomrule
\end{tabular}
\caption{A direct comparison between biomedical models and their respective base models \ctext[RGB]{171,217,217}{Llama-2-(7B/70B)}, \ctext[RGB]{255,210,170}{Mistral-7B-v0.1}, \ctext[RGB]{176,180,251}{Meta-Llama-3-(8B/70B)} and \ctext[RGB]{223,217,174}{Meta-Llama-3.1-(8B/70B)}. The scores show the difference between each model before and after domain adaptation. MCQA shows the reported performance difference averaged over (MedMCQA \citep{medmcqa}, MedQA \citep{medqa} and PubMedQA \citep{jin-etal-2019-pubmedqa}) while Level 1 and 2 show the differences on CLUE. }
\label{tab:benchmark_diff}
\end{table}

Table \ref{tab:overall_results} presents average results for each task, while Table \ref{tab:benchmark_diff} summarizes the relative performance differences between biomedical models and their base models compared to previous MCQA evaluations. Only BioMistral-7B-DARE shows a consistent performance advantage across all six tasks. In contrast, 11 models show performance losses in at least one task, and four biomedical models exhibit declines on all tasks, indicating that domain-specific fine-tuning can harm general task performance.

\begin{figure}[ht]
  \centering
  \includegraphics[width=\columnwidth]{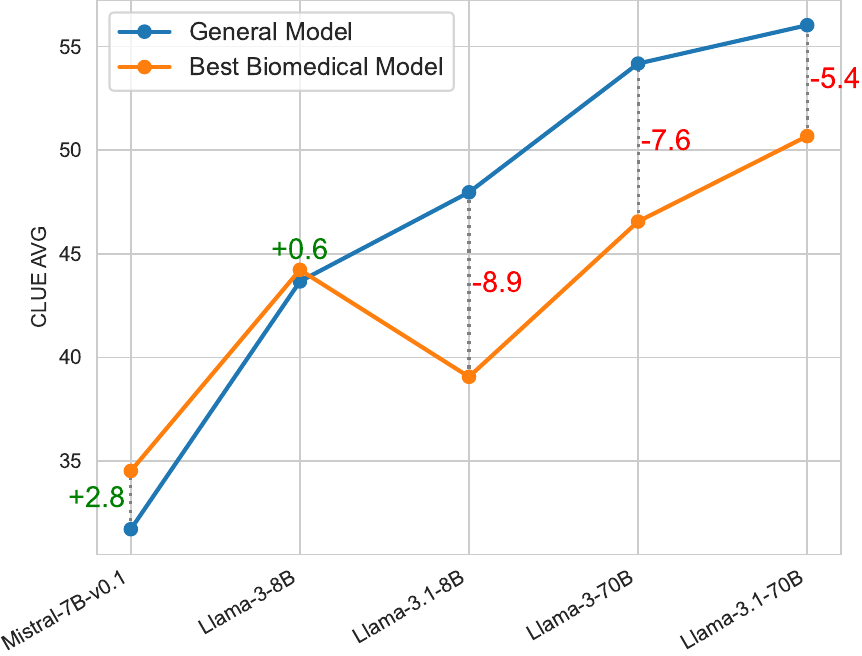}
  \caption{Comparison of average scores between general-domain models and highest scoring biomedical models.}
  \label{average_results}
\end{figure}

Most performance gains are observed in models based on Llama-2 and Mistral-7B-v0.1, while models derived from more recent LLMs frequently underperform after adaptation. Additionally, improvements are more common in models with up to 8B parameters, whereas larger models tend to lose performance after biomedical training. Figure \ref{average_results} shows a comparison between the best-performing biomedical models and their base models. We find slight performance gains for Mistral-7B-v0.1 but clear performance losses for models based on better-performing general-domain LLMs. 

Task complexity also plays a key role: gains are mainly seen in Level 1 tasks, while performance on more complex Level 2 tasks often declines. This suggests biomedical models may struggle with tasks requiring language understanding and reasoning.

Unlike previous reports of biomedical LLM improvements on MCQA evaluations, only two models show slight average gains on both Level 1 and Level 2 tasks on CLUE (see Table \ref{tab:benchmark_diff}).

Overall, general-domain LLMs remain strongest, with Llama3.1-70B emerging as the top performer. Although Llama3-Med42-8B slightly outperforms its base model on simple tasks (+0.56\%), it shows a large drop on Level 2 tasks (-8.03\%).

% Overall, most models exhibit reduced performance following biomedical domain adaptation. Despite improvements in MCQA evaluation, 10 out of 12 models perform worse than their base counterparts on CLUE, with performance degradation being more pronounced in the more complex and realistic level 2 tasks. This suggests while biomedical models do well on simple tasks that test their medical knowledge, they struggle with more complex tasks involving clinical documents, which require deeper language comprehension skills.

% General-domain LLMs demonstrate the highest overall performance, with Llama3.1-70B emerging as the best-performing model. Although Llama3-Med42-8B slightly outperforms Llama-3.1-8B on simpler Level 1 tasks (+0.56\%), it lags significantly on Level 2 tasks (-8.03\%).

% Interestingly, the only biomedical models exhibiting performance gains in both levels—BioMistral-7B and BioMistral-7B-DARE—are based on Mistral-7B-v0.1, the weakest general-domain LLM. In contrast, Mistral-7B-v0.2, a more recent model utilizing the same architecture, achieves substantially better results.

\subsection{Error Analysis}

Primary contributors to biomedical model performance drops are LongHealth task 3 and MeDiSumCode valid code scores (Table \ref{tab:hallucination_tasks}). Biomedical Mistral-7B-based models improve, whereas Llama3-based models show performance decreases of up to $79.15\%$.

LongHealth task 3 measures how often a model correctly returns no answer when information is absent, reflecting hallucination rates. Similarly, MeDiSumCode's valid code scores reveal ICD-10 code fabrication, with low-scoring models incrementing numbers instead of predicting valid codes (see Appendix \ref{appendix:error_analysis}). Notably, Meta-Llama-3-8B-Instruct scored $56.25$ on LongHealth task 3, whereas Llama3-OpenBioLLM-8B dropped to $1.55$. Llama3-OpenBioLLM-70B also underperforms compared to Meta-Llama-3-70B-Instruct.

Beyond hallucinations, biomedical models often fall into repetition loops, generating the same tokens repeatedly and producing incoherent outputs. Additionally, models struggle with instruction adherence, particularly in long-input tasks like LongHealth. This supports previous similar observations \citep{ceballos-arroyo-etal-2024-open}.

\section{Discussion}

Performance declines are observed across various training methods, except for BioMistral-DARE, which uses weight merging, indicating a potential mitigation strategy. However, the superior performance of Mistral-7B-Instruct-v0.2 (Table \ref{tab:overall_results}) suggests that improved general-domain training has a more significant impact than biomedical training.

Many SFT models used generated data, suggesting data quality affects performance. Internist.ai 7b, trained on high-quality data, performed best on Level 2 tasks, reinforcing this hypothesis. 

Improvements were almost exclusive to the lower-performing Mistral-7B-Instruct-v0.1 models, suggesting that recent general models like Llama-3 and Mistral-7B-v0.2 already address these gaps. Tables \ref{tab:hallucination_tasks} and \ref{tab:benchmark_diff} further support this.

\begin{table}[]
\centering
\small
\begin{tabular}{lll}
                         & \multicolumn{1}{l}{LH Task3} & \multicolumn{1}{l}{Valid Codes} \\
\toprule
\cellcolor{orange!35}BioMistral-7B            & {\color[HTML]{02b802} +4.15}                                  & {\color[HTML]{02b802} +17.26}                                 \\
\cellcolor{orange!35}BioMistral-7B-DARE       & {\color[HTML]{02b802} +0.95}                                   & {\color[HTML]{02b802} +18.79}                               \\

\cellcolor{orange!35}Internist.ai 7b & {\color[HTML]{02b802} +45.55}                                 & {\color[HTML]{02b802} +16.32}\\
\cellcolor{blue!30}OpenBioLLM-8B     & {\color[HTML]{C20000} -40.05}                               & {\color[HTML]{C20000} -10.77}                                \\
\cellcolor{blue!30}Med42-8B          & {\color[HTML]{C20000} -12.7}                                & {\color[HTML]{C20000} -6.8}                                  \\
\cellcolor{blue!30}Aloe-8B-Alpha     & {\color[HTML]{C20000} -22.55}                               & {\color[HTML]{C20000} -17.09}                                \\
\cellcolor{blue!30}OpenBioLLM-70B    & {\color[HTML]{C20000} -28.80}                               & {\color[HTML]{C20000} -20.29}                                \\
\cellcolor{blue!30}Med42-70B         & {\color[HTML]{C20000} -79.15}                               & {\color[HTML]{C20000} -15.39} \\
\cellcolor{olive!30}Meditron3-8B         & {\color[HTML]{C20000} -52.15}                               & {\color[HTML]{C20000} -49.19} \\  
\cellcolor{olive!30}Meditron3-70B         & {\color[HTML]{C20000} -54.6}                               & {\color[HTML]{C20000} -4.76} \\
\bottomrule
\end{tabular}
\caption{\ctext[RGB]{255,210,170}{Mistral-7B-v0.1 }, \ctext[RGB]{176,180,251}{Meta-Llama-3-(8B/70B)} and \ctext[RGB]{223,217,174}{Meta-Llama-3.1-(8B/70B)} based models on LongHealth task 3 and percentage of valid ICD-10 codes in MeDiSumCode}
\label{tab:hallucination_tasks}
\end{table}

\section{Conclusion}
Our study suggests that biomedical LLMs are not competing effectively with general-domain models on clinical tasks. While some biomedical models have shown improvements, more recent and larger models are underperforming. Fine-tuning these models with domain-specific data often leads to reduced performance, introducing hallucinations and decreased model stability. This stands in contrast to traditional MCQA evaluations, where biomedical models have previously demonstrated superior performance. Our evaluation provides a more practical assessment of LLM capabilities in real-world healthcare settings. To support further progress in this field, we open-source our evaluation scripts, allowing for broader validation and replication of our results.

\section*{Limitations}

Our study has several limitations that should be considered. Due to the significant computational resources required to run LLMs with up to 141 billion parameters, we did not explore the impact of various model configurations, such as temperature settings, or advanced techniques like chain-of-thought prompting on model performance. Future research should investigate these aspects to gain a more comprehensive understanding of their effects. Additionally, the datasets we use are publicly available resources. As such, we cannot completely prevent data contamination. This limitation underscores the need for future research into robust methods for mitigating data contamination, which is crucial for ensuring the validity of any public LLM benchmark. While we presented novel insights in this paper, their application to clinical data requires further investigation. Future work should refine these methods to enhance their applicability and reliability in clinical settings. Furthermore, our evaluation primarily focused on tasks involving clinical documents and their relevance, but it was not conducted in a realistic clinical setting. Therefore, extensive evaluation through prospective clinical trials is necessary to meet the required safety levels before applying these models to clinical environments.

\bibliography{custom}

\begin{thebibliography}{34}
\providecommand{\natexlab}[1]{#1}

\bibitem[{Abdin et~al.(2024)Abdin, Jacobs, Awan, Aneja, Awadallah, Awadalla, Bach, Bahree, Bakhtiari, Behl et~al.}]{abdin2024phi}
Marah Abdin, Sam~Ade Jacobs, Ammar~Ahmad Awan, Jyoti Aneja, Ahmed Awadallah, Hany Awadalla, Nguyen Bach, Amit Bahree, Arash Bakhtiari, Harkirat Behl, et~al. 2024.
\newblock Phi-3 technical report: A highly capable language model locally on your phone.
\newblock \emph{arXiv preprint arXiv:2404.14219}.

\bibitem[{Adams et~al.(2024)Adams, Busch, Han, Excoffier, Ortala, L{\"o}ser, Aerts, Kather, Truhn, and Bressem}]{adamsLongHealthQuestionAnswering2024}
Lisa Adams, Felix Busch, Tianyu Han, Jean-Baptiste Excoffier, Matthieu Ortala, Alexander L{\"o}ser, Hugo~JWL Aerts, Jakob~Nikolas Kather, Daniel Truhn, and Keno Bressem. 2024.
\newblock \href {https://arxiv.org/abs/2401.14490} {{{LongHealth}}: {{A Question Answering Benchmark}} with {{Long Clinical Documents}}}.
\newblock \emph{Preprint}, arxiv:2401.14490.

\bibitem[{Alsentzer et~al.(2019)Alsentzer, Murphy, Boag, Weng, Jindi, Naumann, and McDermott}]{alsentzer-etal-2019-publicly}
Emily Alsentzer, John Murphy, William Boag, Wei-Hung Weng, Di~Jindi, Tristan Naumann, and Matthew McDermott. 2019.
\newblock \href {https://doi.org/10.18653/v1/W19-1909} {Publicly available clinical {BERT} embeddings}.
\newblock In \emph{Proceedings of the 2nd Clinical Natural Language Processing Workshop}, pages 72--78, Minneapolis, Minnesota, USA. Association for Computational Linguistics.

\bibitem[{Ankit~Pal(2024)}]{OpenBioLLMs}
Malaikannan~Sankarasubbu Ankit~Pal. 2024.
\newblock Openbiollms: Advancing open-source large language models for healthcare and life sciences.
\newblock \url{https://huggingface.co/aaditya/OpenBioLLM-Llama3-70B}.

\bibitem[{{Ben Abacha} and Demner-Fushman(2019)}]{MeQSum}
Asma {Ben Abacha} and Dina Demner-Fushman. 2019.
\newblock On the summarization of consumer health questions.
\newblock In \emph{Proceedings of the 57th Annual Meeting of the Association for Computational Linguistics, ACL 2019, Florence, Italy, July 28th - August 2}.

\bibitem[{Bodenreider(2004)}]{bodenreider2004unified}
Olivier Bodenreider. 2004.
\newblock The unified medical language system (umls): integrating biomedical terminology.
\newblock \emph{Nucleic acids research}, 32(suppl\_1):D267--D270.

\bibitem[{Ceballos-Arroyo et~al.(2024)Ceballos-Arroyo, Munnangi, Sun, Zhang, McInerney, Wallace, and Amir}]{ceballos-arroyo-etal-2024-open}
Alberto~Mario Ceballos-Arroyo, Monica Munnangi, Jiuding Sun, Karen Zhang, Jered McInerney, Byron~C. Wallace, and Silvio Amir. 2024.
\newblock \href {https://doi.org/10.18653/v1/2024.bionlp-1.5} {Open (clinical) {LLM}s are sensitive to instruction phrasings}.
\newblock In \emph{Proceedings of the 23rd Workshop on Biomedical Natural Language Processing}, pages 50--71, Bangkok, Thailand. Association for Computational Linguistics.

\bibitem[{Chen et~al.(2023)Chen, Cano, Romanou, Bonnet, Matoba, Salvi, Pagliardini, Fan, K{\"o}pf, Mohtashami, Sallinen, Sakhaeirad, Swamy, Krawczuk, Bayazit, Marmet, Montariol, Hartley, Jaggi, and Bosselut}]{chenMEDITRON70BScalingMedical2023}
Zeming Chen, Alejandro~Hern{\'a}ndez Cano, Angelika Romanou, Antoine Bonnet, Kyle Matoba, Francesco Salvi, Matteo Pagliardini, Simin Fan, Andreas K{\"o}pf, Amirkeivan Mohtashami, Alexandre Sallinen, Alireza Sakhaeirad, Vinitra Swamy, Igor Krawczuk, Deniz Bayazit, Axel Marmet, Syrielle Montariol, Mary-Anne Hartley, Martin Jaggi, and Antoine Bosselut. 2023.
\newblock \href {https://arxiv.org/abs/2311.16079} {{{MEDITRON-70B}}: {{Scaling Medical Pretraining}} for {{Large Language Models}}}.
\newblock \emph{Preprint}, arxiv:2311.16079.

\bibitem[{Christophe et~al.(2024)Christophe, Kanithi, Munjal, Raha, Hayat, Rajan, Al-Mahrooqi, Gupta, Salman, Gosal, Kanakiya, Chen, Vassilieva, Amor, Pimentel, and Khan}]{christophe2024med42}
Clément Christophe, Praveen~K Kanithi, Prateek Munjal, Tathagata Raha, Nasir Hayat, Ronnie Rajan, Ahmed Al-Mahrooqi, Avani Gupta, Muhammad~Umar Salman, Gurpreet Gosal, Bhargav Kanakiya, Charles Chen, Natalia Vassilieva, Boulbaba~Ben Amor, Marco~AF Pimentel, and Shadab Khan. 2024.
\newblock \href {https://arxiv.org/abs/2404.14779} {Med42 -- evaluating fine-tuning strategies for medical llms: Full-parameter vs. parameter-efficient approaches}.

\bibitem[{Dada et~al.(2025)Dada, Koras, Bauer, Butler, Smith, Kleesiek, and Friedrich}]{dada2025medisumqa}
Amin Dada, Osman~Alperen Koras, Marie Bauer, Amanda Butler, Kaleb~E Smith, Jens Kleesiek, and Julian Friedrich. 2025.
\newblock Medisumqa: Patient-oriented question-answer generation from discharge letters.
\newblock \emph{arXiv preprint arXiv:2502.03298}.

\bibitem[{Gao et~al.(2022)Gao, Dligach, Miller, Xu, Churpek, and Afshar}]{gaoSummarizingPatientsProblems2022}
Yanjun Gao, Dmitriy Dligach, Timothy Miller, Dongfang Xu, Matthew M.~M. Churpek, and Majid Afshar. 2022.
\newblock Summarizing {{Patients}}' {{Problems}} from {{Hospital Progress Notes Using Pre-trained Sequence-to-Sequence Models}}.
\newblock In \emph{Proceedings of the 29th {{International Conference}} on {{Computational Linguistics}}}, pages 2979--2991, Gyeongju, Republic of Korea. International Committee on Computational Linguistics.

\bibitem[{Griot et~al.(2024)Griot, Hemptinne, Vanderdonckt, and Yuksel}]{internist}
Maxime Griot, Coralie Hemptinne, Jean Vanderdonckt, and Demet Yuksel. 2024.
\newblock \href {https://doi.org/10.1093/jamia/ocae120} {{Impact of high-quality, mixed-domain data on the performance of medical language models}}.
\newblock \emph{Journal of the American Medical Informatics Association}, page ocae120.

\bibitem[{Gururajan et~al.(2024)Gururajan, Lopez-Cuena, Bayarri-Planas, Tormos, Hinjos, Bernabeu-Perez, Arias-Duart, Martin-Torres, Urcelay-Ganzabal, Gonzalez-Mallo, Alvarez-Napagao, Ayguadé-Parra, and Garcia-Gasulla}]{gururajan2024aloe}
Ashwin~Kumar Gururajan, Enrique Lopez-Cuena, Jordi Bayarri-Planas, Adrian Tormos, Daniel Hinjos, Pablo Bernabeu-Perez, Anna Arias-Duart, Pablo~Agustin Martin-Torres, Lucia Urcelay-Ganzabal, Marta Gonzalez-Mallo, Sergio Alvarez-Napagao, Eduard Ayguadé-Parra, and Ulises Cortés~Dario Garcia-Gasulla. 2024.
\newblock \href {https://arxiv.org/abs/2405.01886} {Aloe: A family of fine-tuned open healthcare llms}.
\newblock \emph{Preprint}, arXiv:2405.01886.

\bibitem[{Jeong et~al.(2024)Jeong, Garg, Lipton, and Oberst}]{jeong-etal-2024-medical}
Daniel~P Jeong, Saurabh Garg, Zachary~Chase Lipton, and Michael Oberst. 2024.
\newblock \href {https://doi.org/10.18653/v1/2024.emnlp-main.677} {Medical adaptation of large language and vision-language models: Are we making progress?}
\newblock In \emph{Proceedings of the 2024 Conference on Empirical Methods in Natural Language Processing}, pages 12143--12170, Miami, Florida, USA. Association for Computational Linguistics.

\bibitem[{Jiang et~al.(2023)Jiang, Sablayrolles, Mensch, Bamford, Chaplot, Casas, Bressand, Lengyel, Lample, Saulnier et~al.}]{jiang2023mistral}
Albert~Q Jiang, Alexandre Sablayrolles, Arthur Mensch, Chris Bamford, Devendra~Singh Chaplot, Diego de~las Casas, Florian Bressand, Gianna Lengyel, Guillaume Lample, Lucile Saulnier, et~al. 2023.
\newblock Mistral 7b.
\newblock \emph{arXiv preprint arXiv:2310.06825}.

\bibitem[{Jiang et~al.(2024)Jiang, Sablayrolles, Roux, Mensch, Savary, Bamford, Chaplot, de~las Casas, Hanna, Bressand, Lengyel, Bour, Lample, Lavaud, Saulnier, Lachaux, Stock, Subramanian, Yang, Antoniak, Scao, Gervet, Lavril, Wang, Lacroix, and Sayed}]{jiang2024mixtral}
Albert~Q. Jiang, Alexandre Sablayrolles, Antoine Roux, Arthur Mensch, Blanche Savary, Chris Bamford, Devendra~Singh Chaplot, Diego de~las Casas, Emma~Bou Hanna, Florian Bressand, Gianna Lengyel, Guillaume Bour, Guillaume Lample, Lélio~Renard Lavaud, Lucile Saulnier, Marie-Anne Lachaux, Pierre Stock, Sandeep Subramanian, Sophia Yang, Szymon Antoniak, Teven~Le Scao, Théophile Gervet, Thibaut Lavril, Thomas Wang, Timothée Lacroix, and William~El Sayed. 2024.
\newblock \href {https://arxiv.org/abs/2401.04088} {Mixtral of experts}.
\newblock \emph{Preprint}, arXiv:2401.04088.

\bibitem[{Jin et~al.(2021)Jin, Pan, Oufattole, Weng, Fang, and Szolovits}]{medqa}
Di~Jin, Eileen Pan, Nassim Oufattole, Wei-Hung Weng, Hanyi Fang, and Peter Szolovits. 2021.
\newblock What disease does this patient have? a large-scale open domain question answering dataset from medical exams.
\newblock \emph{Applied Sciences}, 11(14):6421.

\bibitem[{Jin et~al.(2019)Jin, Dhingra, Liu, Cohen, and Lu}]{jin-etal-2019-pubmedqa}
Qiao Jin, Bhuwan Dhingra, Zhengping Liu, William Cohen, and Xinghua Lu. 2019.
\newblock \href {https://doi.org/10.18653/v1/D19-1259} {{P}ub{M}ed{QA}: A dataset for biomedical research question answering}.
\newblock In \emph{Proceedings of the 2019 Conference on Empirical Methods in Natural Language Processing and the 9th International Joint Conference on Natural Language Processing (EMNLP-IJCNLP)}, pages 2567--2577, Hong Kong, China. Association for Computational Linguistics.

\bibitem[{Johnson et~al.(2021)Johnson, Bulgarelli, Pollard, Horng, Celi, and Mark}]{mimiciv_v1}
Alistair Johnson, Lucas Bulgarelli, Tom Pollard, Steven Horng, Leo~Anthony Celi, and Roger Mark. 2021.
\newblock \href {https://doi.org/10.13026/s6n6-xd98} {Mimic-iv}.

\bibitem[{Johnson et~al.(2016)Johnson, Pollard, Shen, Lehman, Feng, Ghassemi, Moody, Szolovits, Anthony~Celi, and Mark}]{johnson2016mimic}
Alistair~EW Johnson, Tom~J Pollard, Lu~Shen, Li-wei~H Lehman, Mengling Feng, Mohammad Ghassemi, Benjamin Moody, Peter Szolovits, Leo Anthony~Celi, and Roger~G Mark. 2016.
\newblock Mimic-iii, a freely accessible critical care database.
\newblock \emph{Scientific data}, 3(1):1--9.

\bibitem[{Labrak et~al.(2024)Labrak, Bazoge, Morin, Gourraud, Rouvier, and Dufour}]{labrakBioMistralCollectionOpenSource2024a}
Yanis Labrak, Adrien Bazoge, Emmanuel Morin, Pierre-Antoine Gourraud, Mickael Rouvier, and Richard Dufour. 2024.
\newblock \href {https://arxiv.org/abs/2402.10373} {{{BioMistral}}: {{A Collection}} of {{Open-Source Pretrained Large Language Models}} for {{Medical Domains}}}.
\newblock \emph{Preprint}, arxiv:2402.10373.

\bibitem[{Lin(2004)}]{lin-2004-rouge}
Chin-Yew Lin. 2004.
\newblock \href {https://aclanthology.org/W04-1013} {{ROUGE}: A package for automatic evaluation of summaries}.
\newblock In \emph{Text Summarization Branches Out}, pages 74--81, Barcelona, Spain. Association for Computational Linguistics.

\bibitem[{Luo et~al.(2023)Luo, Zhang, Fan, Yang, Wu, Qiao, and Nie}]{luoBioMedGPTOpenMultimodal2023}
Yizhen Luo, Jiahuan Zhang, Siqi Fan, Kai Yang, Yushuai Wu, Mu~Qiao, and Zaiqing Nie. 2023.
\newblock \href {http://arxiv.org/abs/2308.09442} {{BioMedGPT}: {Open} {Multimodal} {Generative} {Pre}-trained {Transformer} for {BioMedicine}}.
\newblock \emph{arXiv preprint}.
\newblock ArXiv:2308.09442 [cs] version: 2.

\bibitem[{Moor et~al.(2023)Moor, Banerjee, Abad, Krumholz, Leskovec, Topol, and Rajpurkar}]{moor2023foundation}
Michael Moor, Oishi Banerjee, Zahra Shakeri~Hossein Abad, Harlan~M Krumholz, Jure Leskovec, Eric~J Topol, and Pranav Rajpurkar. 2023.
\newblock Foundation models for generalist medical artificial intelligence.
\newblock \emph{Nature}, 616(7956):259--265.

\bibitem[{Neumann et~al.(2019)Neumann, King, Beltagy, and Ammar}]{neumann-etal-2019-scispacy}
Mark Neumann, Daniel King, Iz~Beltagy, and Waleed Ammar. 2019.
\newblock \href {https://doi.org/10.18653/v1/W19-5034} {{S}cispa{C}y: {F}ast and {R}obust {M}odels for {B}iomedical {N}atural {L}anguage {P}rocessing}.
\newblock In \emph{Proceedings of the 18th BioNLP Workshop and Shared Task}, pages 319--327, Florence, Italy. Association for Computational Linguistics.

\bibitem[{OpenMeditron(2024)}]{meditron3_modelcard}
OpenMeditron. 2024.
\newblock \href {https://huggingface.co/OpenMeditron} {Meditron3 model card}.

\bibitem[{Organization(2004)}]{ICD}
World~Health Organization. 2004.
\newblock Icd-10 : international statistical classification of diseases and related health problems : tenth revision.

\bibitem[{Pal et~al.(2022)Pal, Umapathi, and Sankarasubbu}]{medmcqa}
Ankit Pal, Logesh~Kumar Umapathi, and Malaikannan Sankarasubbu. 2022.
\newblock \href {https://proceedings.mlr.press/v174/pal22a.html} {Medmcqa: A large-scale multi-subject multi-choice dataset for medical domain question answering}.
\newblock In \emph{Proceedings of the Conference on Health, Inference, and Learning}, volume 174 of \emph{Proceedings of Machine Learning Research}, pages 248--260. PMLR.

\bibitem[{Romanov and Shivade(2018)}]{DBLP:journals/corr/abs-1808-06752}
Alexey Romanov and Chaitanya Shivade. 2018.
\newblock \href {https://arxiv.org/abs/1808.06752} {Lessons from natural language inference in the clinical domain}.
\newblock \emph{CoRR}, abs/1808.06752.

\bibitem[{Saab et~al.(2024)Saab, Tu, Weng, Tanno, Stutz, Wulczyn, Zhang, Strother, Park, Vedadi et~al.}]{saab2024capabilities}
Khaled Saab, Tao Tu, Wei-Hung Weng, Ryutaro Tanno, David Stutz, Ellery Wulczyn, Fan Zhang, Tim Strother, Chunjong Park, Elahe Vedadi, et~al. 2024.
\newblock Capabilities of gemini models in medicine.
\newblock \emph{arXiv preprint arXiv:2404.18416}.

\bibitem[{Singhal et~al.(2023)Singhal, Azizi, Tu, Mahdavi, Wei, Chung, Scales, Tanwani, Cole-Lewis, Pfohl et~al.}]{singhal2023large}
Karan Singhal, Shekoofeh Azizi, Tao Tu, S~Sara Mahdavi, Jason Wei, Hyung~Won Chung, Nathan Scales, Ajay Tanwani, Heather Cole-Lewis, Stephen Pfohl, et~al. 2023.
\newblock Large language models encode clinical knowledge.
\newblock \emph{Nature}, 620(7972):172--180.

\bibitem[{Tunstall et~al.(2023)Tunstall, Beeching, Lambert, Rajani, Rasul, Belkada, Huang, von Werra, Fourrier, Habib, Sarrazin, Sanseviero, Rush, and Wolf}]{tunstall2023zephyr}
Lewis Tunstall, Edward Beeching, Nathan Lambert, Nazneen Rajani, Kashif Rasul, Younes Belkada, Shengyi Huang, Leandro von Werra, Clémentine Fourrier, Nathan Habib, Nathan Sarrazin, Omar Sanseviero, Alexander~M. Rush, and Thomas Wolf. 2023.
\newblock \href {https://arxiv.org/abs/2310.16944} {Zephyr: Direct distillation of lm alignment}.
\newblock \emph{Preprint}, arXiv:2310.16944.

\bibitem[{Weed(1964)}]{Weed1964}
Lawrence~L. Weed. 1964.
\newblock \href {https://doi.org/10.1007/bf02945791} {Medical records, patient care, and medical education}.
\newblock \emph{Irish Journal of Medical Science}, 39(6):271–282.

\bibitem[{Zhang et~al.(2019)Zhang, Kishore, Wu, Weinberger, and Artzi}]{zhang2019bertscore}
Tianyi Zhang, Varsha Kishore, Felix Wu, Kilian~Q Weinberger, and Yoav Artzi. 2019.
\newblock Bertscore: Evaluating text generation with bert.
\newblock In \emph{International Conference on Learning Representations}.

\end{thebibliography}

\appendix

\section{Task Details}

\subsection{Metrics}
\label{appendix:metrics}
For open-ended tasks, we report the F1-score between the model predictions and ground truth unigrams (ROUGE-1), bigram (ROUGE-2), and the longest common subsequence (ROUGE-L)\footnote{https://huggingface.co/spaces/evaluate-metric/rouge} \citep{lin-2004-rouge}. We compute the BERTScore \citep{zhang2019bertscore} on clinical documents to measure semantic similarity using an encoder trained on MIMIC III\footnote{emilyalsentzer/Bio\_ClinicalBERT} \citep{alsentzer-etal-2019-publicly}. We first tuned the score rescaling baselines for MIMIC IV discharge summaries. For Problem Summaries and MeDiSumQA, we also extract the Unified Medical Language System (UMLS) \citep{bodenreider2004unified} entities with scispacy \citep{neumann-etal-2019-scispacy} and compute their F1-score to consider medical abbreviations and synonyms. 
When evaluating MedDiSumCode, we calculate the ratio of valid ICD-10 codes. We use the python package icd10-cm\footnote{https://pypi.org/project/icd10-cm/} to probe the validity of ICD-10 codes.  We distinguish between exact match (EM) and the match of the first three characters of the codes, which is an approximate match (AP) based on the hierarchical structure of ICD-10 codes.

\begin{table*}
\centering
\begin{tabular}{lll}\toprule
Model Name &Base Model &Type of Training \\\toprule
Meditron-7B &Llama2-7B &Continued pretraining \\
Internist.ai 7B &Mistral-7B-v0.1 &Continued pretraining + SFT \\
BioMistral-7B &Mistral-7B-Instruct-v0.1 &Continued pretraining \\
BioMistral-7B-DARE &Mistral-7B-Instruct-v0.1 &Continued pretraining +DARE \\
Llama3-OpenBioLLM-8B &Meta-Llama-3-8B-Instruct &SFT + DPO \\
Llama3-Med42-8B & Meta-Llama-3-8B-Instruct & SFT + DPO \\
Llama3-Aloe-8B-Alpha &Meta-Llama-3-8B-Instruct &SFT + DPO \\
Meditron3-8B &Meta-Llama-3.1-8B-Instruct & - \\
Meditron-70B &Llama-2-70B &Continued pretraining \\
Llama3-OpenBioLLM-70B &Meta-Llama-3-70B-Instruct &SFT + DPO \\
Llama3-Med42-70B & Meta-Llama-3-8B-Instruct &SFT + DPO \\
Meditron3-70B &Meta-Llama-3.1-70B-Instruct & - \\

\bottomrule
\end{tabular}
\caption{Evaluated Biomedical Models}
\label{biomed_models}

\end{table*}

\section{Experimental setup}

\subsection{Computational Resources}

All experiments were conducted on an NVIDIA DGX A100 640GB node with 8x NVIDIA A100 80GB Tensor Core GPUs within three days, resulting in approximately 1536 GPU hours.

\subsection{Models}
\label{appendix:models}
Table \ref{biomed_models} lists all biomedical models we evaluated. 

\subsection{Prompting}
\label{appendix:prompting}

We apply few-shot prompting and use the instruction template on Hugging Face for the instruction-tuned models. For the other models, we concatenate the system prompt, few-shot examples, and user prompt into one string separated by double newlines. For the level one evaluation, we performed 3-shot prompting. For level two, we provide one shot with the exception of LongHealth, where we provide no examples due to the content length.

\begin{figure*}[h]
  \centering
  \includegraphics[width=\textwidth]{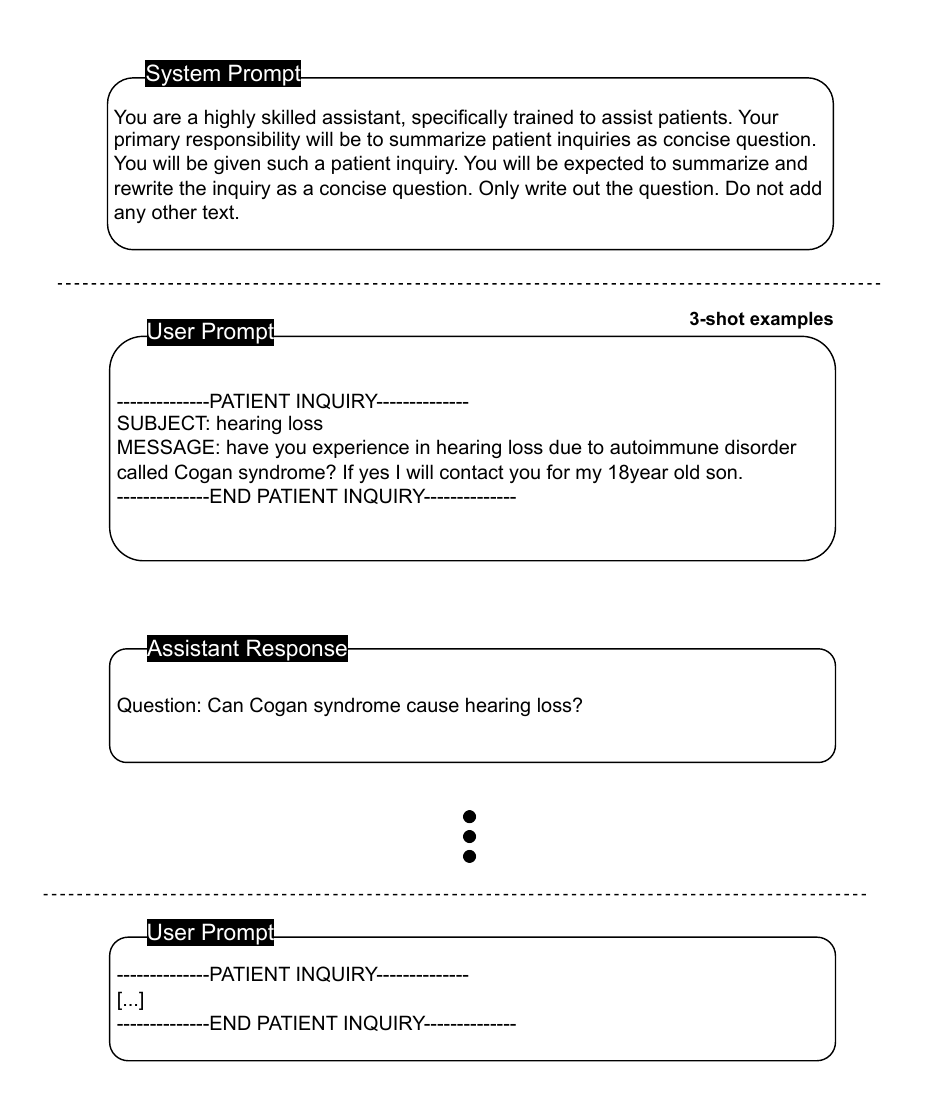}
  \caption{MeQSum prompt format with example.}
  \label{meqsum_example_prompts}
\end{figure*}

\begin{figure*}[h]
  \centering
  \includegraphics[width=\textwidth]{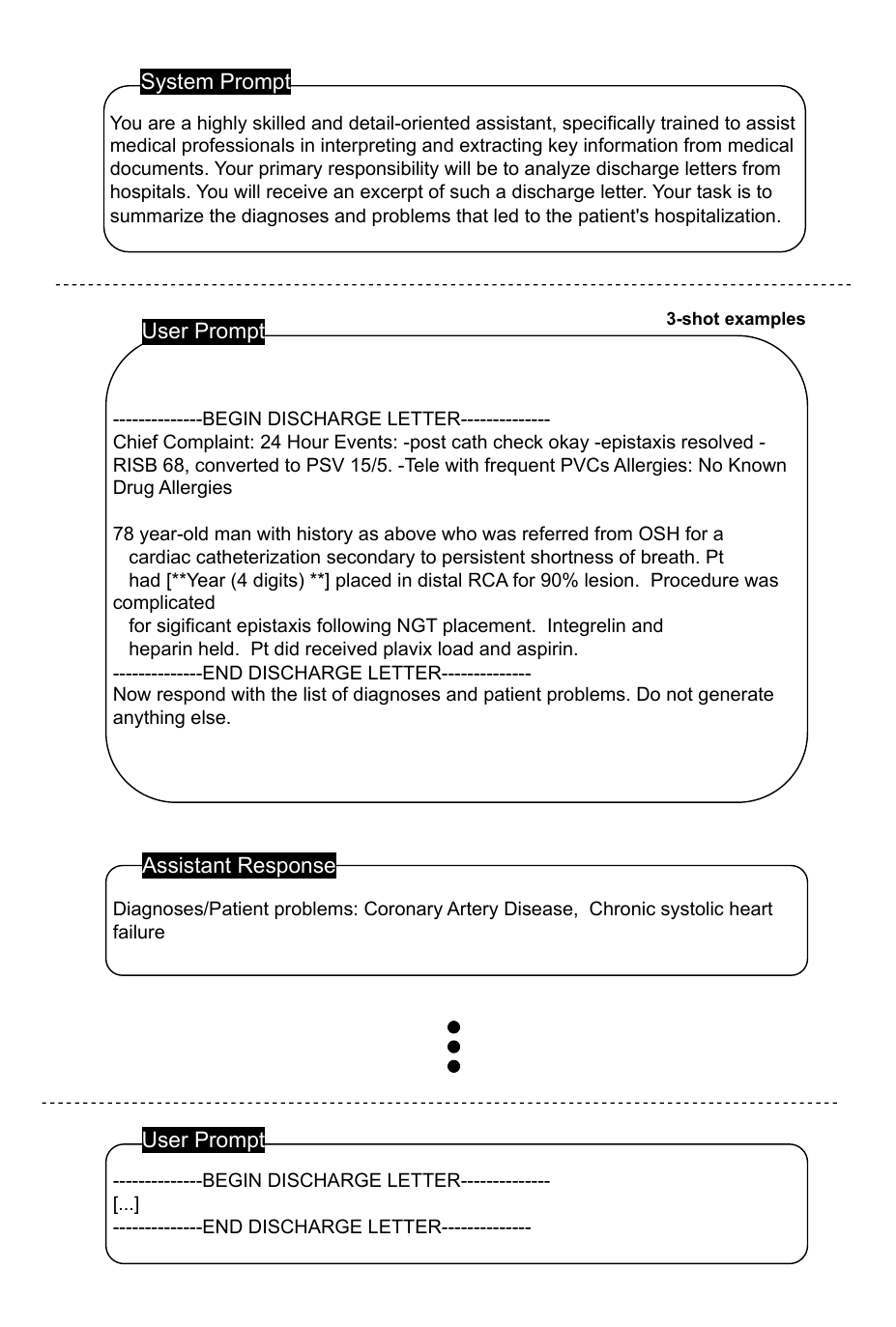}
  \caption{Problem Summary prompt format with example.}
  \label{probsum_example_prompts}
\end{figure*}

\begin{figure*}[h]
  \centering
  \includegraphics[width=\textwidth]{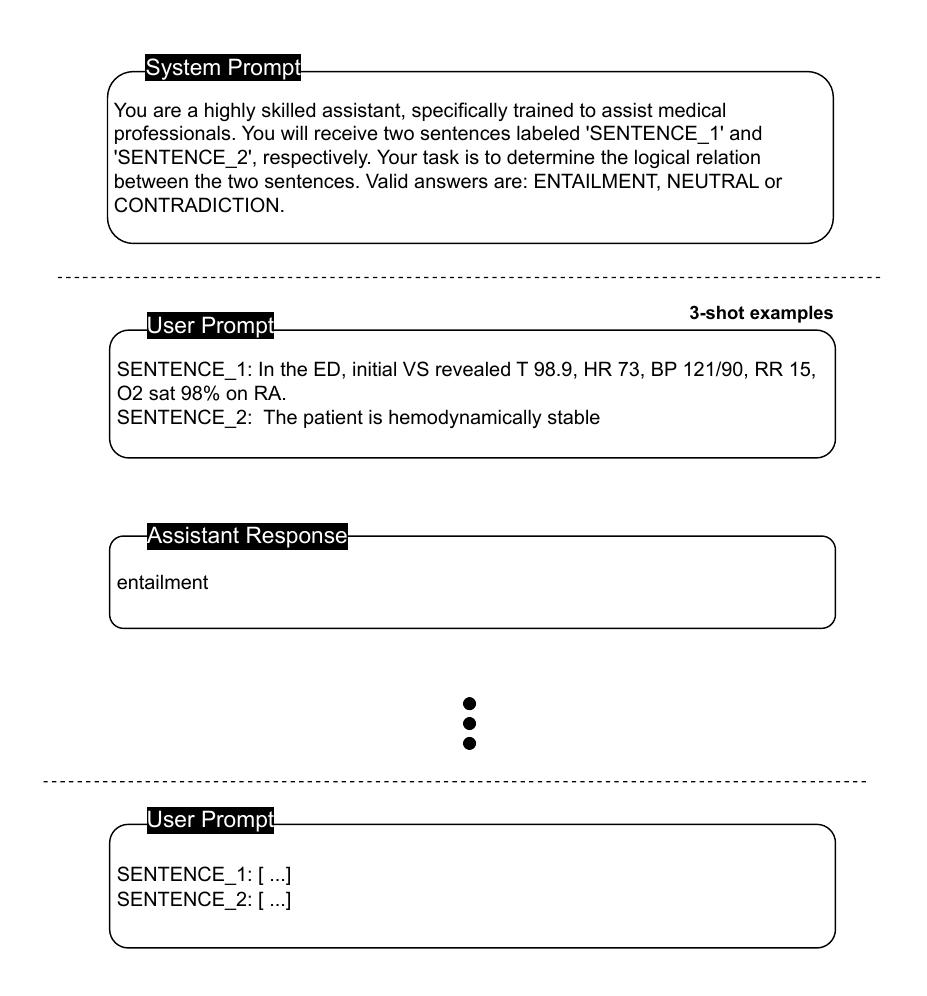}
  \caption{MedNLI prompt format with example.}
  \label{mednli_example_prompts}
\end{figure*}

\begin{figure*}[h]
  \centering
  \includegraphics[width=0.9\textwidth]{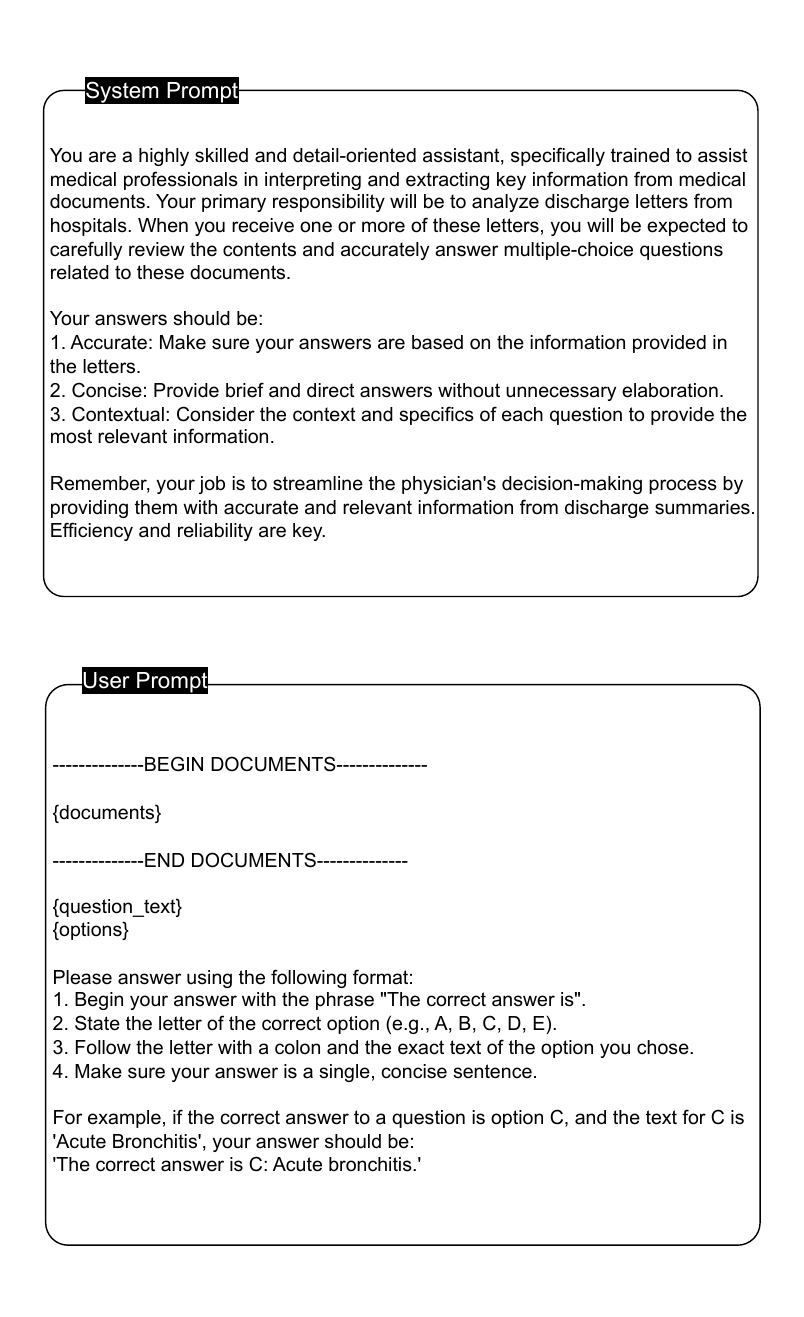}
  \caption{LongHealth prompt format.}
  \label{longhealth_example_prompts}
\end{figure*}

\begin{figure*}[h]
  \centering
  \includegraphics[width=0.9\textwidth]{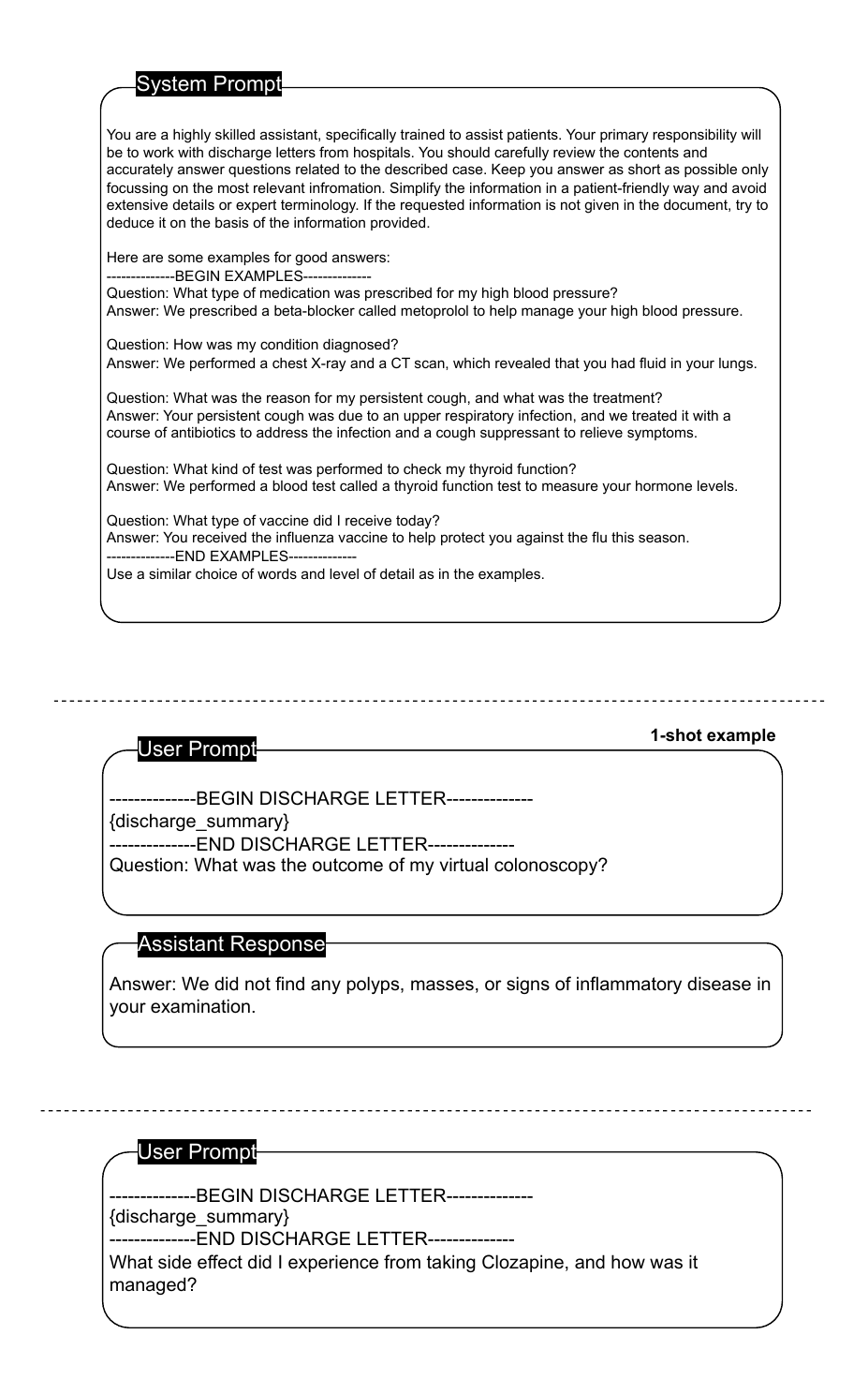}
  \vspace{-2cm}
  \caption{MeDiSumQA prompt format.}
  \label{medisumqa_example_prompts}
\end{figure*}

\begin{figure*}[h]
  \centering
  \includegraphics[width=\textwidth]{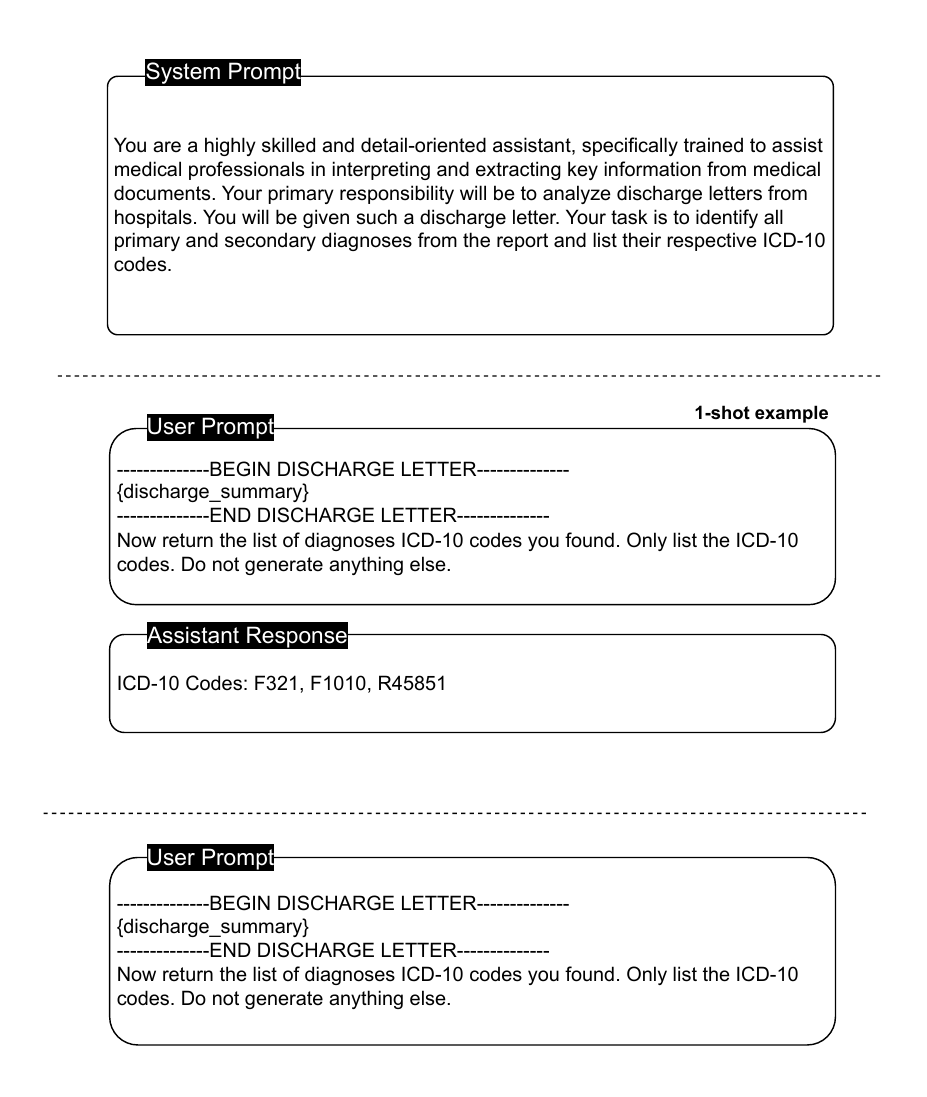}
  \caption{MeDiSumCode prompt format.}
  \label{medisumcode_example_prompts}
\end{figure*}

Figures \ref{meqsum_example_prompts}, \ref{probsum_example_prompts}, \ref{mednli_example_prompts}, \ref{longhealth_example_prompts}, \ref{medisumqa_example_prompts}, and
\ref{medisumcode_example_prompts} are showing the prompt formats we are using for the different benchmark tasks. If the input length allowed this, we also included sample texts from the datasets.

% \section{Results}
% \label{appendix:results}
% Table \ref{tab:main_results2} shows the detailed benchmark results for all models.

\begin{figure*}[h]
  \centering
  \includegraphics[width=\textwidth]{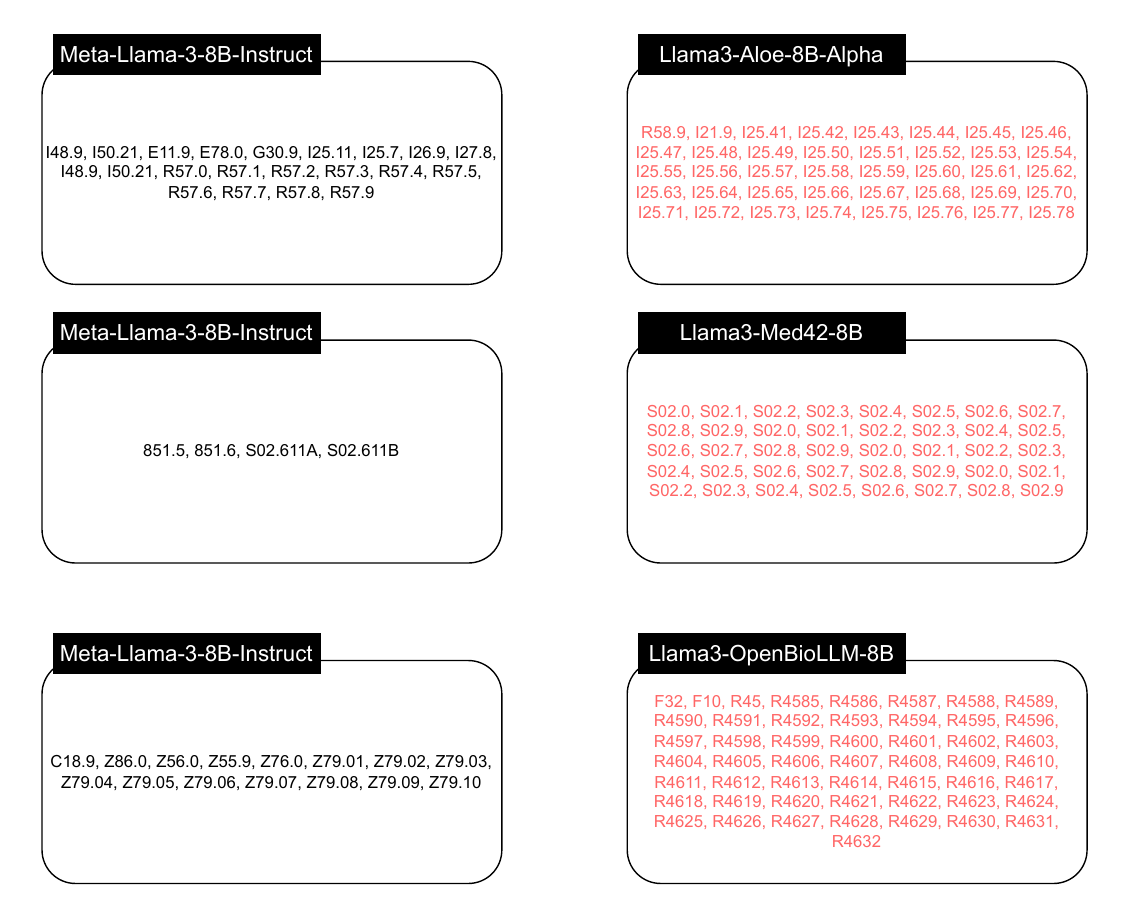}
  
  \caption{Biomedical models that show the described counting behavior compared to their base model.}
  \label{fig:counting_example}
\end{figure*}

\subsection{Error Analysis}
\label{appendix:error_analysis}
Figure \ref{fig:counting_example} shows some examples of the described type of error with regard to counting.

\end{document}